\pdfoutput=1
\documentclass[PHM, 2020]{PHMSociety}

\usepackage[utf8]{inputenc}
\usepackage[english]{babel}

\usepackage{amsthm}
\usepackage{graphicx}
\usepackage{amsmath}
\theoremstyle{definition}
\newtheorem{definition}{Definition}[section]

\begin{document}

\title{A Relearning Approach to Reinforcement Learning for control of Smart Buildings}

\author{%
	Avisek Naug\authorNumber{1}, Marcos Quiñones -Grueiro \authorNumber{2}, and Gautam Biswas\authorNumber{3}
}

\address{
	\affiliation{{1,2,3 }}{Vanderbilt University, Nashville, TN, USA}{ 
		{\email{avisek.naug@vanderbilt.edu}}\\ 
		{\email{marcos.quinones@vanderbilt.edu}}\\ 
		{\email{gautam.biswas@vanderbilt.edu}}
		} 
}

\maketitle

\phmLicenseFootnote{Avisek Naug}

\begin{abstract}
This paper demonstrates that continual relearning of control policies using incremental deep reinforcement learning (RL) can improve policy learning for non-stationary processes. This approach has been demonstrated in a data-driven “smart building environment” that we use as a test-bed for developing HVAC controllers for reducing energy consumption of large buildings on our university campus. 
The non-stationarity in building operations and weather patterns makes it imperative to develop control strategies that are adaptive to changing conditions. On-policy RL algorithms, such as Proximal Policy Optimization (PPO) represent an approach for addressing this non-stationarity, but they cannot be applied to safety-critical systems. As an alternative, we develop an incremental RL technique that simultaneously reduces building energy consumption without sacrificing overall comfort. We compare the performance of our incremental RL controller to that of a  static RL controller that does not implement the relearning function. The performance of the static controller diminishes significantly over time, but the relearning controller adjusts to changing conditions while ensuring comfort and optimal energy performance.
\end{abstract}

\section{Introduction}\label{Introduction}
Energy efficient control of Heating, Ventilation and Air Conditioning (HVAC) systems is an important aspect of building operations because they account for the major share of energy consumed by buildings. 
Most large office buildings,which are significant energy consumers, are structures with complex, internal energy flow dynamics and complex interactions with their environment. Therefore, building energy management is a difficult problem. Traditional building energy control systems are based on heuristic rules to control the parameters of the building's HVAC systems. 
However, analysis of historical data shows that such rule-based heuristic control is inefficient because the rules are based on simplified assumptions about weather and building operating conditions. 

Recently, there has been a lot of research on \textit{smart buildings} with smart controllers that sense the building state and environmental conditions to adjust the HVAC parameters to optimize building energy consumption \cite{shaikh2014review}. Model Predictive Control (MPC) methods have been successfully deployed for smart control \cite{maasoumy2014handling}, but traditional MPC methods require accurate models to achieve good performance.  Developing such models for large buildings may be an intractable problem~\cite{smarra2018data}. Recently, Data-driven \textit{MPC} based on random forest methods have been used to solve demand-response problems for moderate size buildings \cite{smarra2018data}, but is not clear how they may scale up for continuous control of large buildings. 

Reinforcement Learning (RL) methods have recently gained traction for controlling energy consumption and comfort in smart buildings because they provide several advantages. Unlike MPC methods for robust receding horizon control \cite{wei2017deep}, they have the ability to learn a locally optimal control policy without simulating the system dynamics over long time horizons. Instead, RL methods use concepts from Dynamic Programming to select the optimal actions. A number of reinforcement learning controllers for buildings have been proposed, where the building behavior under different environmental conditions are learnt from historical data \cite{Naug2019}. These approaches are classified as data driven or Deep Reinforcement Learning approaches. \cite{Lillicrap2016}. 


However, current data driven approaches for RL do not take into account the non-stationary behaviors of the building and its environment. Building operations and the environments in which they operate are continually changing, often in unpredictable ways. In such situations, the Deep RL controller performance degrades because the data that was used to train the controller becomes `stale'. The solution to this problem is to detect changes in the building operations and its environment, and relearn the controller using data that is more relevant to the current situation. This paper proposes such an approach, where we relearn the controller at periodic intervals to maintain its relevance, and thus its performance. 

The rest of the paper is organized as follows. Section \ref{LitReview} presents a brief review of some of the current approaches in model and data-driven reinforcement learning, and the concept of non-stationarity in MDPs. 
Section \ref{ContinuousLearning} formally introduces the RL problem for non-stationary systems that we tackle in this paper. Section \ref{Implementation}  then develops our data driven modeling as well as the reinforcement learning schemes for `optimal' building energy management. 
Section \ref{Results} discusses our experimental results, and Section \ref{Conclusion} presents our conclusions and directions for future work.

\section{Literature Review}\label{LitReview}


Traditional methods for developing RL controllers of systems have relied on accurate dynamic models of the system (model-based approaches) or data-driven approaches. We briefly review model-based and data-driven approaches to RL control, and then introduce the notion of non-stationary systems, where traditional methods for RL policy learning are not effective, 

\subsection{Reinforcement Learning with Model Based Simulators}\label{Model Based Simulators}
Typical physics-based models of building energy consumption, use conservation of energy and mass to construct thermodynamic equations to describe system behavior.  
\cite{wei2017deep} applied Deep \textit{Q-Learning} methods \cite{Mnih2015} to optimize the energy consumption and ensure temperature comfort in a building simulated using EnergyPlus\cite{eplus}, a whole building energy simulation program. \cite{Moriyama2018} obtained cooling energy savings of $22\%$ on an EnergyPlus simulated model of a data-center using a natural policy gradient based algorithm called TRPO \cite{Schulman2015}. Similarly, \cite{Li2019} used an off policy algorithm called DDPG \cite{Lillicrap2016} to obtain $11\%$ cooling energy savings in an EnergyPlus simulation of a data-center. To deal with sample inefficiency in \textit{on-policy learning}, \cite{Hosseinloo2020} developed an event-triggered RL approach, where the control action changes when the system crosses a boundary function in the state space. They used a one-room EnergyPlus thermal to demonstrate their approach.

\subsection{Reinforcement Learning with Data Driven Approaches}\label{Data Driven Simulators} \label{DDRL}
The examples above describe RL approaches applied to simple building architectures. As discussed, creating a model based simulator for large, complex buildings can be quite difficult \cite{Park,Kim2011}. Alternatively, more realistic approaches for RL applied to large buildings rely on historical data from the building to learn data-driven models or directly use the data as experiences from which a policy is learnt. 
\cite{Nagy2018} developed simulators from data-driven models and then used them for finite horizon control. \cite{naug2018data} used Support Vector Regression to develop a building energy consumption model, and then used stochastic gradient methods to optimize energy consumption. \cite{Costanzo2016} used value-based neural networks to learn the thermodynamics model of a building. The energy models were then optimized using \textit{Q-learning} \cite{sutton2018reinforcement}. Subsequently, \cite{Naug2019} used a \textit{DDPG} \cite{Lillicrap2016} approach with a sampling buffer to develop a policy function that minimized energy consumption without sacrificing comfort.  Anther recent approach that has successfully applied deep RL to data-driven building energy optimization includes \cite{mocanu2018line}.

\subsection{Non Stationary MDPs}\label{litreviewnsmdp}
The data-driven approaches presented in Section \ref{DDRL} do not address the non-stationarity of the large buildings. Non-stationary behaviors can be attributed to multiple sources. For example, weather patterns, though seasonal, can change abruptly in unexpected ways. Similarly, conditions in a building can change quickly, e.g., when a large number of people enter the building for an event, or components of the HVAC system, degrade of fail, e..g, stuck valves, or failed pumps. When such situations occur, a RL controller, trained on the past experiences, cannot adapt to the unexpected changes in the system and environment, and, therefore, performs sub-optimally. Some work \cite{Mankowitz2018,tamar2014scaling,Shashua2017} has been proposed to address non-stationarity in the environments by improving  the \textit{value function} under the worst case conditions\cite{Iyengar2005} of the non-stationarity. 

Other approaches try to minimize a \textit{regret function} instead of finding the optimal policy for \textit{non-stationary MDPs}. The regret function measures the sum of missed rewards when we compare the state value from a start state between current best policy and the  target policy in hindsight \textit{i.e.}, they tell us what actions would have been appropriate after the episode ends. This regret is then optimized to get better actions. \cite{Hallak2015} applied this approach to context-driven MDPs (each context may represent a different non-stationary behavior)  to find the piecewise stationary optimal policies for each context. They proposed a clustering algorithm to find a set of contexts. \cite{Jaksch,Gajane2019} also minimize the regret based on an average reward formulation instead of a state value function. \cite{Padakandla2019} proposed a non-stationary MDP control method under a model-free setting by using a context detection method proposed in \cite{Prabhu20}. These approaches assume knowledge of a known set of possible environment models beforehand, which may not be possible in real systems.  Moreover, they are model-based, i.e., they assume the MDP models are available. Therefore, they cannot be applied in a model free setting. 


To address non-stationarity issues in complex buildings we extend previous research in this domain to make the following contributions to data-driven modeling and RL based control of buildings:

\begin{itemize}
    \item We retrain the dynamic behavior models of the building and its environment at regular intervals to ensure that the models respond to the distributional shifts in the system behavior, and, therefore, provide an accurate representation of the behavior.
    \item By not relearning the building and its environment model from scratch, we ensure the repeated training is not time consuming. This also has the benefit of the model not being susceptible to the catastrophic forgetting \cite{Kirkpatrick2017} of the past behavior which is common in neural networks used for online training and relearning.
    \item We relearn the policy function; i.e., the HVAC controller every time the dynamic model of the system is re learnt,  so that it adapts to the current conditions in the building.
\end{itemize}

In the rest of this paper, we develop the relearning algorithms, and demonstrate the benefits of this incremental relearning approach on the controller efficiency.

\section{Optimal Control with Reinforcement Learning}
\label{Non-StationaryMDP}

Reinforcement learning (RL) represents a class of machine learning methods for solving optimal control problems, where an agent learns by continually interacting with an environment~\cite{sutton2018reinforcement}. In brief, the agent observes the state of the environment, and based on this state/observation takes an action, and notes the reward it receives for the $(state,action)$ pair. The agent's ultimate goal is to compute a \textit{policy}, i.e., a mapping from the environment states to the actions that maximizes the expected sum of reward. RL has been cast as a stochastic optimization method for solving Markov Decision Processes (MDPs), when the MDP is not known. We define RL problem more formally below.

\begin{definition}[Markov Decision Process]
A Markov decision process is defined by a four tuple: $M=\{\mathit{S},\mathit{A},\mathit{T},\mathit{R}\},$ where $S$ represents the set of possible states in the environment. The transition function $\mathit{T}: \mathit{S} \times \mathit{A} \times \mathit{S} \rightarrow [0, 1]$ defines the probability of reaching state $s'$ at $t+1$ given that action $a \in \mathit{A}$ was chosen in state $s \in \mathit{S}$ at \textit{decision epoch} $t$, $T=p(s'|s,a)= Pr\{s_{t+1}=s'|s_t=s,a_t=a\}$. The reward function $R: \mathit{S} \times \mathit{A} \rightarrow \Re$ estimates the immediate reward $\mathit{R}\sim r(s,a)$ obtained from choosing action $a$ in state $s$.
\end{definition}

The objective of the agent is to find a policy $\pi^*$ that maximizes the accumulated discounted rewards it receives over the future. The optimization criteria is the following:
\begin{equation}
\label{opt}
V^{\pi^*}(s)= \max_{\pi \in \Pi} V^{\pi}(s) \; , \; \forall s \in \mathit{S},
\end{equation}
where $V^{\pi}: \mathit{S} \rightarrow \mathit{R}$ is called value function and it is defined as 
\begin{equation}
\label{vf1}
V^{\pi}(s) = E\left[ \sum_{t=0}^{\infty} \gamma^{t}\mathit{R}(s_t,a_t)|s_0=s\right] \; , \; \forall s \in \mathit{S},
\end{equation}
where $0 < \gamma \leq 1$ is called the discount factor, and it determines the weight assigned to future rewards. In other words, the weight associated with future rewards decays with time.

An optimal deterministic Markovian policy satisfying Equation \ref{opt} exists if the following conditions are satisfied
\begin{enumerate}
\item $|\mathit{R}\sim r(s,a)| \leq C < \infty, \forall a \in \mathit{A}, s \in \mathit{S}$
\item $\mathit{T}$ and $\mathit{R}$ do not change over time. 
\end{enumerate}

If a MDP satisfies the second condition, it is called a \textit{stationary} MDP. However, most real world systems undergo changes that cause their dynamic model, represented by the transition function $\mathit{T}$, to change over time \cite{Dulac2019}. In other words, these systems exhibit \textit{non stationary behaviors}. Non stationary behaviors may happen because the components of a system degrade, and/or the environment in which a system operates changes, causing the models that govern the system behavior to change over time. In case of large buildings, the weather conditions can change abruptly, or changes in occupancy or faults in building components can cause unexpected and unanticipated changes in the system's behavior model. In other words, $\mathit{T}$ is no longer invariant, but it may change over time. Therefore, a more realistic model of the interactions between an agent and its environment is defined by a non stationary MDP (NMDP) \cite{Puterman2014}.

\begin{definition}[Non-Stationary Markov Decision Process]
A non-stationary Markov decision process is defined by a 5-tuple: $M=\{\mathit{S},\mathit{A},\mathcal{T},(p_t)_{t \in \mathcal{T}},(r_t)_{t \in \mathcal{T}}\}$. $S$ represents the set of possible states that the environment can reach at decision epoch $t$. $\mathcal{T}=\{1,2,...,N\}$ is the set of decision epochs with $N \leq +\infty$. $\mathit{A}$ is the action space. $p_t(s'|s,a)$ and $r_t(s,a)$ represent the transition function and the reward function at decision epoch $t$, respectively. 
\end{definition}

In the most general case, the optimal policy for a NMDP, $\pi_t$ is also non stationary. The value of state $s$ at decision epoch $t$ within an infinite horizon NMDP is defined for a stochastic policy as follows:
\begin{equation}
\label{vf}
V^{\pi}_t(s) = E\left[ \sum_{i=t}^{\infty} \gamma^{i-t}\mathit{R_i}(s_i,a_i)|s_t=s,\, a_i \sim \pi_i,\, s_{i+1} \sim p_i \right] \; 
\end{equation}

Learning optimal policies from non-stationary MDPs is particularly difficult for non-episodic tasks when the agent is unable to explore the time axis at will. However, real systems do not change arbitrarily fast over time. Hence, we can assume that changes occur slowly over time. This assumption is know as the \textit{regularity hypothesis} and it can be formalized by using the notion of Lipschitz Continuity (LC) applied to the transition and reward functions of a non-stationary MDP \cite{Lecarpentier2019}. This results in the definition of Lipschitz Continuous NMDP (LC-NMDP)

\begin{definition}[($L_p,L_r$) -LC-NMDP]
An ($L_p,L_r$) -LC-NMDP is a NMDP whose transition and reward functions are respectively $L_p$-LC and $L_r$-LC w.r.t. time, i.e. $\forall (t,\hat{t},s,s^{'},a)$ \\
$W_1(p_t(.|s,a),p_{\hat{t}}(.|s,a)) \leq L_p|t-\hat{t}|$ and \\ $|r_t(s,a,s^{'})-r_{\hat{t}}(s,a,s^{'})| \leq L_r|t-\hat{t}|$ \\
where $W_1$ represents the Wasserstein distance and it is used to quantify the distance between two distributions.
\end{definition}

Although learning from the true NMDP is generally not possible because the agent does not have access to the true NSMDP model, it is possible to learn a quasi-optimal policy from interacting with temporal slices of the NMDP assuming the LC-property. This means that the agent can learn using a stationary MDP of the environment at time $t$. Therefore, the trajectory generated by a  LC-NMDP $\{s_0,r_0,...,s_k\}$ is assumed to be generated by a sequence of stationary MDPs $\{MDP_{t_0},...,MDP_{t_0+k-1}\}$. In the next section, we present a continuous learning approach for optimal control of non stationary processes based on this idea.

\section{Continual Learning Approach for Optimal Control of Non-Stationary Systems}\label{ContinuousLearning}
\begin{figure}
  \centering
  \includegraphics[width=\columnwidth,height=5.3cm]{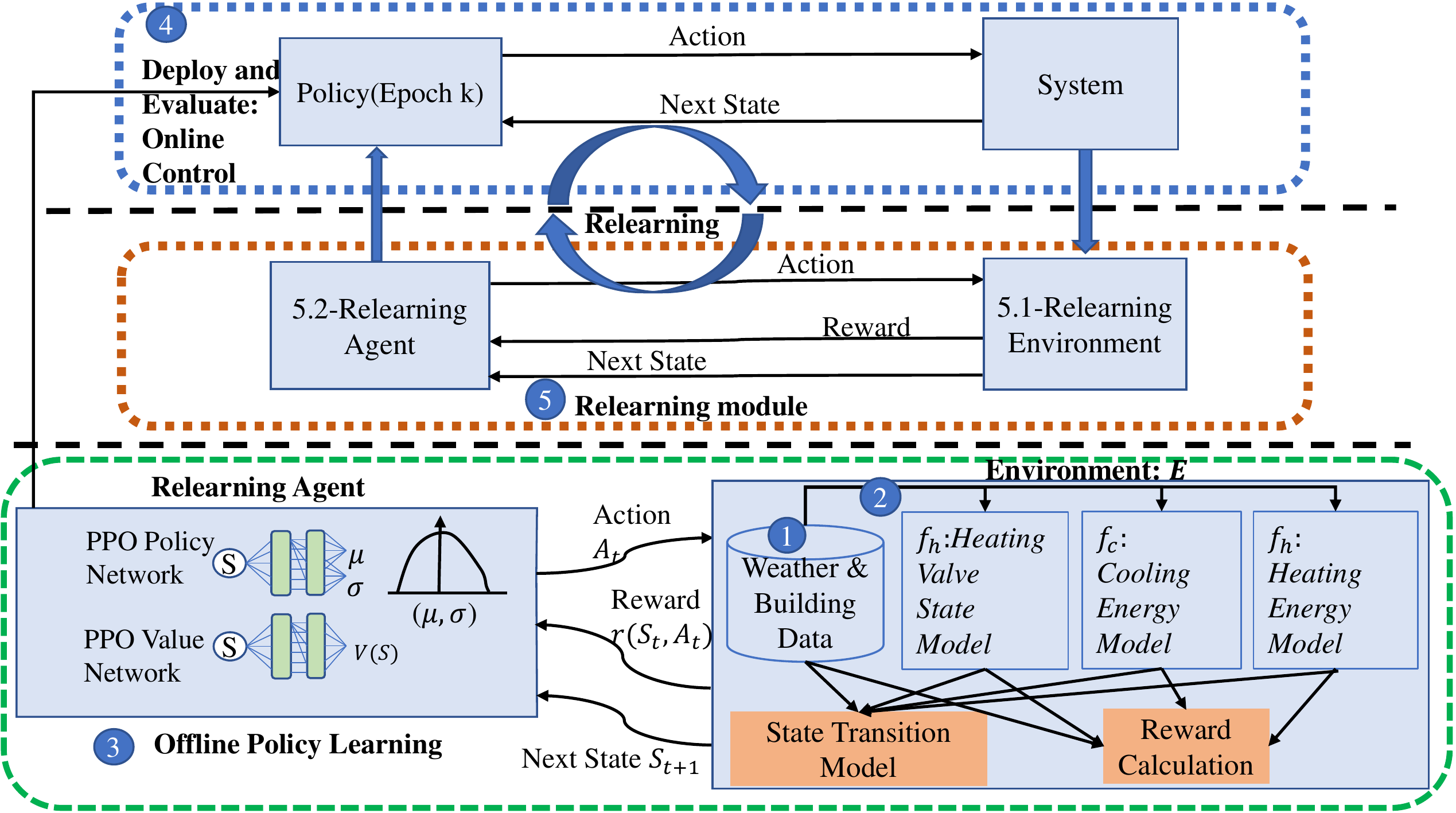}
  \caption{Schematic of our Proposed Approach}
  \label{fig:approach}
\end{figure}

The proposed approach has two main steps: an initial offline learning process followed by continual learning process. Figure \ref{fig:approach} presents the proposed approach organized in the following steps which are annotateed as 1, 2 $\ldots$ in the figure:

\begin{itemize}
    \item Step 1. \textit{Data collection}. Typically this represents historical data that may be available about system operations. In our work, we start with a data set containing information on past weather conditions and the building's energy-related variables. This data set may be representative of one or more operating conditions of the non stationary system, in our case, the building, 
    \item Step 2. \textit{Deriving a dynamic model of the environment}. In our case, this is the building energy consumption model, given relevant building and weather parameters.
    \begin{itemize}
        \item A state transition model is defined in terms of state variables (inputs and outputs) and the dynamics of the system are learned from the data set.  
        \item The reward function used to train the agent is defined. 
    \end{itemize}
    \item Step 3. \textit{Learning an initial policy}. A policy is learned offline by interacting with the environment model derived in the previous step. 
    \item Step 4. \textit{Deployment}. The policy learned is deployed online, i.e., in the real environment, and experiences from theses interaction are collected. 
    \item Step 5. \textit{Relearning}. In general, the relearning module would be invoked based on some predefined performance parameters, for example, when average accumulated reward value over small intervals of time is monotonically decreasing. When this happens:
    \begin{itemize}
        \item the transition model of the environment is updated based on the recent experiences collected from the interaction with the up-to-date policy.
        \item The current policy is re-trained offline, much like Step 3, by interacting with the environment now using the updated transition model of the system.
    \end{itemize}
\end{itemize}

We will demonstrate that this method works if the regularity hypothesis is satisfied, i.e., the environment changes occur after sufficiently long intervals, to allow for the offline relearning step (Step 5) to be effectively applied. In this work, we also assume that the reward function, $R$, is stationary, and does not have to be re-derived (or re-learned) when episodic non stationary changes occur in the system.

Another point to note is that our algorithm uses a two-step \textit{off line} process to learn a new policy: (1) learn the dynamic (transition) model of the system from recent experiences; and (2) relearn the policy function using the new transition model of the system. This approach addresses two important problems: (1) policy learning happens off line, therefore, additional safety check and verification methods can be applied to the learned policy before deployment $-$ this is an important consideration for \textit{safety critical} systems; and (2) the relearning process can use an appropriate mix of past experiences and recent experiences to relearn the environment model and the corresponding policy. Thus, it addresses the \textit{catastrophic forgetting} problem discussed earlier. This approach also provides a compromise between \textit{off policy} and \textit{on policy} learning in RL, by addressing to some extent the \textit{sample inefficiency} problem.

We use Long Short-Term Memory (\textit{LSTM}) Neural Network to model the dynamics of the system and the the Proximal Policy Optimization (\textit{PPO}) algorithm to train the control policy. PPO is one of the best known reinforcement learning algorithm for learning optimal control law in short periods of time. Next, we  describe our approach to modeling the dynamic environment using LSTMs, and the reinforcement learning algorithm for learning and relearning the building controllers (i.e., the policy functions). 

\subsection{Long Short-Term Memory Networks for Modeling Dynamic Systems}\label{LSTM}
Despite their known success in machine learning tasks, such as image classification, deep learning approaches for energy consumption prediction have not been sufficiently explored \cite{Amasyali2018}. In recent work, Recurrent neural networks (RNN) have demonstrated their effectiveness for load forecasting when compared against standard Multi Layer Perceptron (MLP) architectures  \cite{Kong2019,Rahman2018}.
\begin{figure}[!ht]
  \centering
  \includegraphics[width=\linewidth,height=4.0cm]{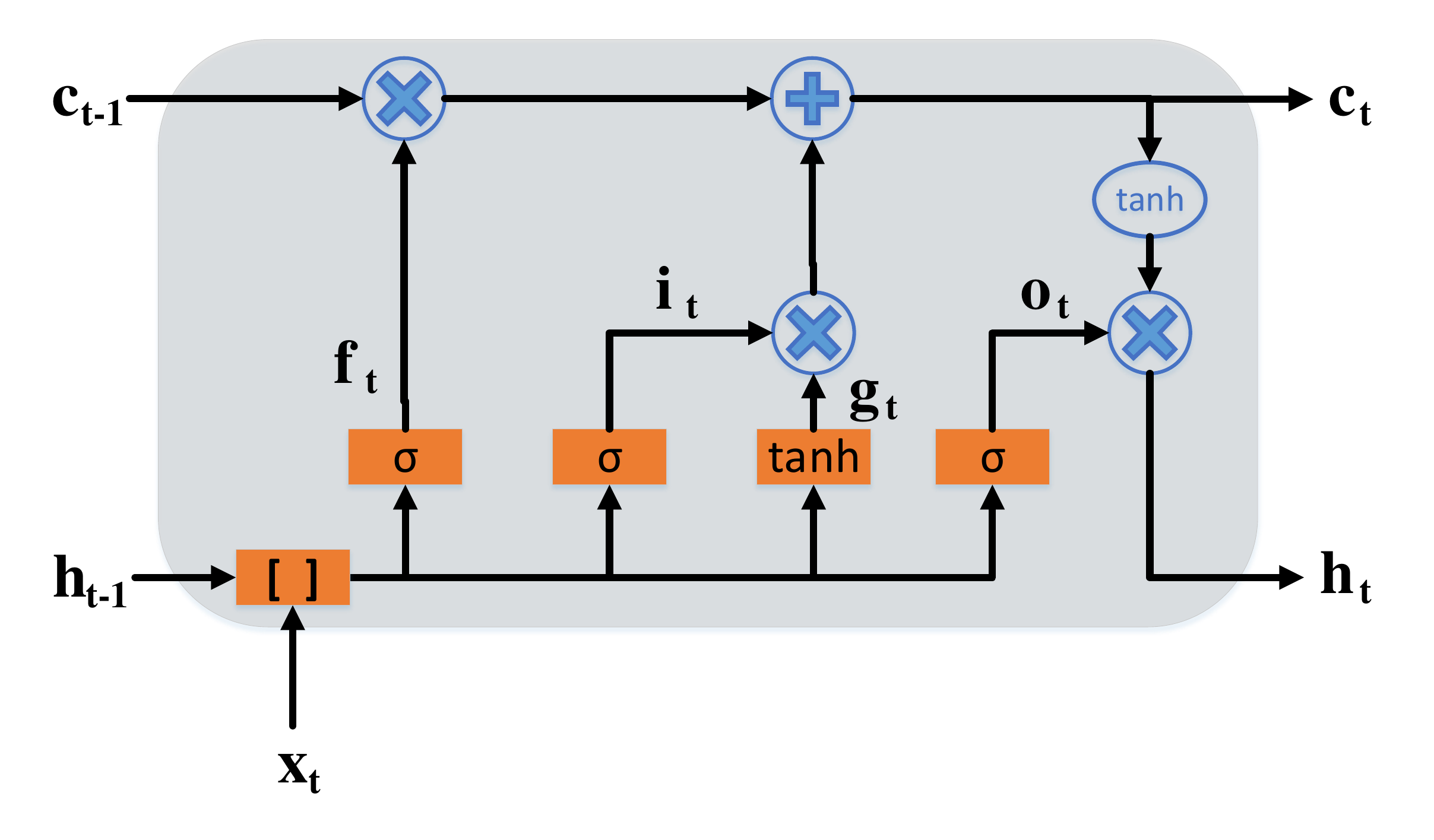}
  \caption{Time unrolled architecture of the basic LSTM neural network block}
  \label{figlstm}
\end{figure}
Among the variety of RNN architectures, Long-Short Term Memory (LSTM) networks have the flexibility for modeling complex dynamic relationships and the capability to overcome the so-called vanishing/exploding gradient problem associated with training the recurrent networks \cite{hochreiter1997long}.  Moreover, LSTMs can capture arbitrary long-term dependencies, which are likely in the context of energy forecasting tasks for large, complex buildings. The architecture of an LSTM model is represented in Figure~\ref{figlstm}. It captures non-linear long-term dependencies among the variables based on the following equations: 

\begin{align}
\mathbf{i_t}=\sigma(W_{ix}\mathbf{x_t}+ W_{ih}\mathbf{h_{t-1}} + \mathbf{b_i}) \label{eq:input}\\
\mathbf{f_t}=\sigma(W_{fx}\mathbf{x_t}+ W_{fh}\mathbf{h_{t-1}} + \mathbf{b_f}) \label{eq:forget}\\
\mathbf{o_t}=\sigma(W_{ox}\mathbf{x_t}+ W_{oh}\mathbf{h_{t-1}} + \mathbf{b_o}) \label{eq:output}\\
\mathbf{g_t}=\phi(W_{gx}\mathbf{x_t}+ W_{gh}\mathbf{h_{t-1}} + \mathbf{b_g}) \label{eq:memcell1}\\
\mathbf{c_t}=\mathbf{g_t}\odot\mathbf{i_t}+\mathbf{c_{t-1}}\odot\mathbf{f_t} \label{eq:memcell2}\\
\mathbf{h_{t}}=\phi(\mathbf{c_t})\odot\mathbf{o_t}, \label{eq:hidden},
\end{align}

where $\mathbf{x_t} \in \Re^{m}$, $\mathbf{h_{t}} \in \Re^{n}$, and $\mathbf{c_{t-1}} \in \Re^{n}$ represent the input variables, hidden state and memory cell state vectors respectively; $\odot$ stands for element-wise multiplication; and $\sigma$ and $\phi$ are the sigmoid and tanh activation functions. 

The adaptive update of values in the input and forget gates ($\mathbf{i_t},\mathbf{f_t}$) provide LSTMs the ability to remember and forget patterns (Equation ~\ref{eq:memcell2}) over time. The information accumulated in the memory cell is transferred to the hidden state scaled by the output gate ($\mathbf{o_t}$). Therefore, training this network consists of learning the input-output relationships for energy forecasting by adjusting the eight weight matrices and bias vectors.

\subsection{Proximal Policy Optimization}\label{PPO}
The Proximal Policy Optimization(\textit{PPO}) algorithm \cite{schulman2017proximal} has its roots in the Natural Policy Gradient method \cite{kakade2002natural}, whose goal was to improve the common issues encountered in the application of policy gradients. Policy gradient methods\cite{sutton2000policy} represent better approaches to creating optimal policies, especially when compared to value-based reinforcement learning techniques.  Value-based methods suffer from convergence issues when used with function approximators (Neural networks). Policy gradient methods also have issues with high variability, which have been addressed by Actor-Critic methods \cite{konda2000actor}. However, choosing the best step-size for policy updates was the single biggest issue that was addressed in \cite{kakade2002approximately}. PPO replaces the log of action probability in the policy gradient equation
\begin{equation*}
    \mathrm{L}^{PG}(\theta) = \hat{\mathrm{E}}_t\Bigg[log\pi_{\theta}(a_t|s_t)\hat{A}_t\Bigg],
\end{equation*}
with the probability ratio $r(\theta) = \frac{\pi_{\theta}(a_t|s_t)}{\pi_{\theta_{old}}(a_t|s_t)}$ inspired by \cite{kakade2002approximately}. Here, the current parameterized control policy is denoted by $\pi_{\theta}(a|s)$. $\hat{A}_t$ denotes the advantage of taking a particular action $a$ compared to the average of all other actions in state $s$. According to the authors of PPO, this addresses the issue of the step size partially as they need to limit the values of this probability ratio. So they modify the objective function further to provide a \textit{Clipped Surrogate Objective} function,
\begin{equation}\label{Lclip}
    \mathrm{L}^{CLIP}(\theta) = \hat{\mathrm{E}}_t\Bigg[min(
    r(\theta)\hat{A}_t,clip(r(\theta),1-\epsilon,1+\epsilon)\hat{A}_t
    )
    \Bigg].
\end{equation}
The best policy is found by maximizing the above objective. The above objective has several interesting properties that makes PPO easily implementable and fast to reach convergence during each optimization step. The clipping ensures that the policy does not update too much in a given direction when the Advantages are positive. Also, when the Advantages are negative, the clipping makes sure that the probability of choosing those actions are not decreased too much. In other words, it strikes a balance between exploration and exploitation with monotonic policy improvement by using the probability ratio. 

Experiments run on the \textit{Mujoco} platform, show that the PPO algorithm outperforms many other state of the art reinforcement learning algorithms~\cite{engstrom2019implementation}. This motivates our use of this algorithm in our relearning approach.

The PPO algorithm implements a parameterized policy $\pi_{\theta}(a|s)$ using a neural network whose input is the state vector $\mathit{S}$ and the output is the mean $\mu$ and standard deviation $\sigma$ of the best possible action in that state. The policy network is trained using the clipped objective function (see Equation \ref{Lclip}) to obtain the best controller policy. 
A second neural network called the value network, $V(\mathit{S})$, keeps track of the values associated with the states under this policy. This is subsequently used to estimate the advantage $\hat{A}_t$ of action $\mathit{A}$ in state $\mathit{S}$. Its input is also $\mathit{S}$ and its output is a scalar value indicating the average return from that state when policy $\pi_{\theta}(a|s)$ is followed. This network is trained using the \textit{TD} error \cite{sutton2018reinforcement}.

\section{Problem Formulation for The Building Environment}\label{System Description and Problem Formulation}
We start with a description of our building environment and formulate the solution of the energy optimization problem by using our continuous RL approach. This section presents the dynamic data-driven model of building energy consumption and the reward function we employ to derive our control policy.

\subsection{System Description}\label{System Description}
\begin{figure}[!ht]
  \includegraphics[width=\columnwidth,height=5.5cm]{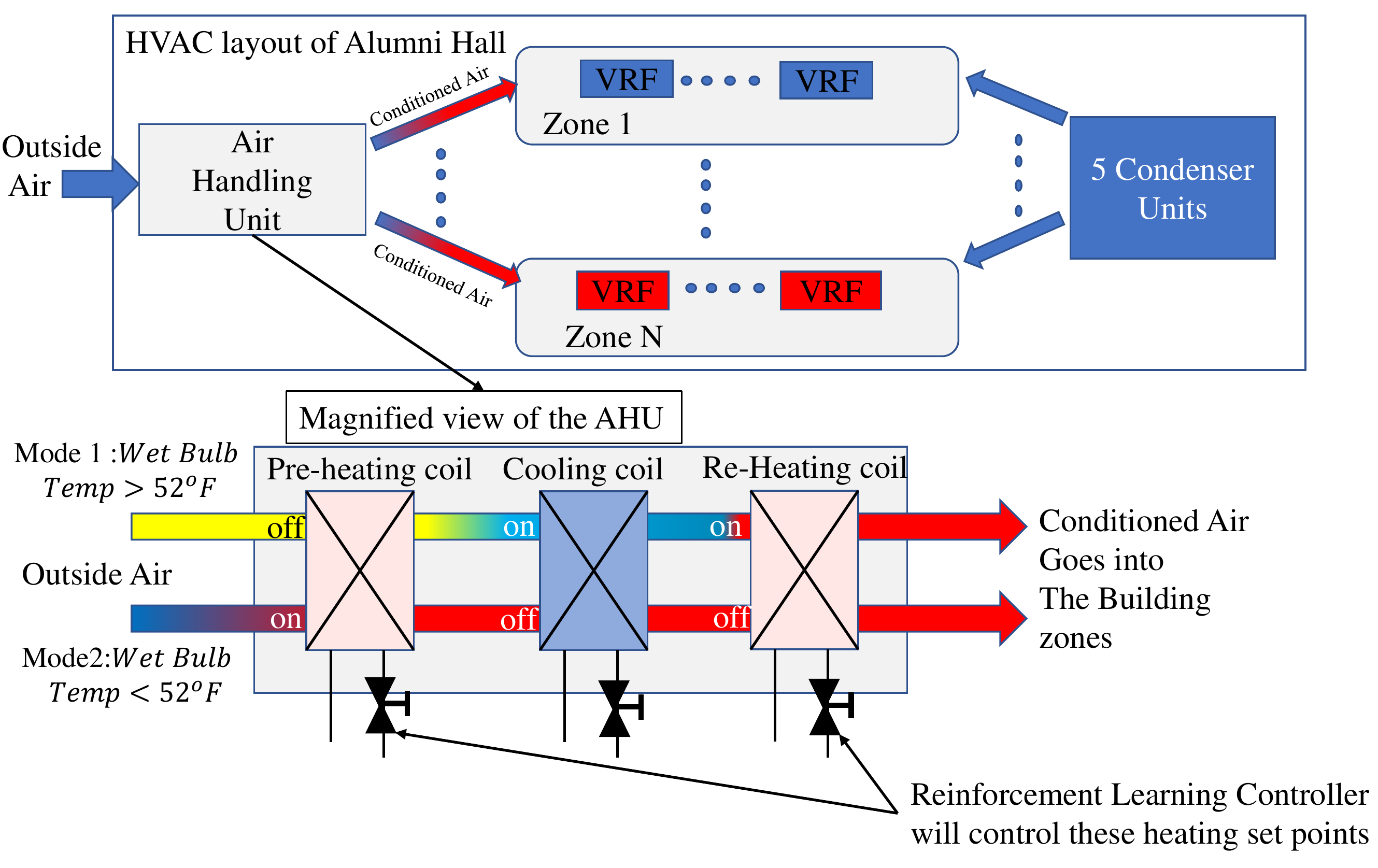}
  \caption{Simplified schematic of the HVAC system under Study}
   \label{fig:alumni_hvac}
\end{figure}

The system under consideration is a large three-storeyed building on our university campus. It has a collection of individual office spaces, classrooms, halls, a gymnasium, a student lounge, and a small cafeteria. The building climate is controlled by a combination of Air Handling Units(\textit{AHU}) and Variable Refrigerant Flow (\textit{VRF}) systems~\cite{Naug2019}. The configuration of the HVAC system is shown in Figure \ref{fig:alumni_hvac}.

The AHU brings in fresh air from the outside and adjusts the air's temperature and humidity before releasing it into the building. Typically, the desired humidity level in the building is set to $50$\%, and the desired temperature values are set by the occupants. Typically, the air is released into the building at a neutral temperature (usually $65^oF$ or $72^oF$). The VRF units in the different zones further heat or cool the air according to the respective temperature set-point (defined by the occupants' preferences). 
The AHU has two operating modes depending on the outside wet bulb temperature. When the wet bulb temperature is above $52^oF$, only the cooling and the reheat coils operate. The AHU dehumidifies the air using the cooling coil to reduce the air temperature to $52^oF$, thus causing a condensation of the excess moisture,  and then heats it back up to a specific value that was originally determined by a rule-based controller (either $65^oF$ or $72^oF$). When the wet bulb temperature is below $52^oF$ (implying the humidity of the outside air is below $50$\%), only the preheat coil operates to heat the incoming cold air to a predefined set-point. The discharge temperature (reheating and preheating set-point depending on the operating mode) will be defined by our RL controller. The appropriate setting of this set-point would allow to reduce the work that must be done by the VRF units, as well as to prevent the building from becoming too cold during cooler weather. 


\subsection{Problem Formulation}\label{Problem Formulation}

The goals of our RL controller is to determine the discharge air temperature set-point of the AHU to minimize the total heating and cooling energy consumed by the building without sacrificing comfort. We will formulate the RL problem by specifying the state-space, the action-space, the reward function, and the transition function for the our building environment.

\subsubsection{State Space}\label{State Space}

The overall energy consumption of our building depends on how the AHU operates but also on exogenous factors such as the weather variability and the building occupancy. The evolution of the weather does not depend on the state of the building. Therefore, the control problem we are trying to solve must be framed as a non-stationary and Exogenous State MDP. The latter can be formalized as follows 
\\

\begin{definition}[Exogenous State Markov Decision Process]
An Exogenous State Markov decision process is defined by a Markov Decision Process which transition function satisfies the following property

\small{
$p(e',x'|e,x,a)= Pr\{x_{t+1}=x'|x_t=x)\}Pr\{e_{t+1}=e'|e_t=e,x_t=x,a_t=a\}$
}\\
where the state space of the MDP is divided into two sub-spaces $\mathit{S} = \xi \times \varepsilon$ such that $x \in \xi$ and $e \in \varepsilon$.
\end{definition}

The above definition can be easily extended to the non-stationary case by considering the time dependency of the transition functions. The condition described above can be interpreted as if there is a subset of state variables whose change is independent from the actions taken by the agent. For our building, the subset of exogenous variables of the subspace $\xi$ are: (1) Outside Air Temperature (\textit{oat}), (2) Outside Air Relative Humidity (\textit{orh}), (3) Wet Bulb Temperature (\textit{wbt}),  (4) Solar irradiance (\textit{sol}), (5) Average Building Temperature Preference Set Point (\textit{avg-stpt}). The remaining variables corresponding to the subspace $\varepsilon$ are (6) AHU Supply Air Temperature (\textit{sat}), (7) Heating energy for the Entire Building($f_h$) and (7) Cooling energy for the Entire Building ($f_c$). Since building occupancy is not measured at this moment, we cannot incorporate that variable to our state space.

\subsubsection{Action Space}\label{Action Space}
The action space $\mathit{A}_t$ of the MDP in each epoch $\mathcal{T}$ is the change in the neutral discharge temperature set-point. As discussed before, the wet bulb temperature determines the AHU operating mode. The valves and actuators that operate the HVAC system have a certain latency in their operation. This means that our controller must not arbitrarily change the discharge temperature set-point. We therefore adopted a safer approach where the action space $\mathit{A}$ is defined as a continuous variable $\in [-2^oF,+2^oF]$ that represents the change with respect to the previous set-point. This means that at every output instant (in the present problem we have set the output to every $30$ minutes), the controller can change at most the discharge temperature set-point by this amount. 

\subsection{Transition Model}
Taking into consideration that the state and action space of the building are continuous, the transition function will comprise 3 components. 

First, the transition function of the exogenous state variables $Pr(x'|x)$ is not explicitly modeled (\textit{oat}, \textit{orh}, \textit{wbt}, \textit{sol}, and \textit{avg-stpt}). Their next state ($\xi_{t+1}$) is determined by looking up at weather database forecasting for the next time step. These variables are available at $5$ minute intervals through a Metasys portal of our building; solar irradiance, $sol$, is available from external data sources. There are no humidity or occupancy sensors inside the building, therefore, we did not consider them as part of the exogenous state variables.

The supply air temperature and the heating and cooling energies are the non-exogenous variables. The change in the supply air temperature \textit{sat} is a function of the current temperature and the set-point selected by the agent.
\begin{equation}\label{sattrans}
    sat_{t+1}=f(sat_t,setpoint_t),
\end{equation}

Here, the controller action $\mathit{A}_t$ will determine what the new set-point $setpoint_t$ will be and subsequently the supply air temperature will approximate that value. We do not create a transition function for this variable since we obtain its value from a sensor installed in the AHU.

Lastly, the heating and cooling energy variables($f_h$ and $f_c$) are determined by the transition functions
\begin{equation}\label{hwetrans}
    f_{h, t+1} = F(\xi_{t+1}, f_{h,t}),
\end{equation}
\begin{equation}\label{cwetrans}
    f_{c, t+1} = F(\xi_{t+1}, f_{c,t}),
\end{equation}
where $\xi_{t+1} = [oat_{t+1},orh_{t+1},wb_{t+1}t,sol_{t+1}, avg-stpt_{t+1},sat_{t+1}]$. As discussed in the last section, we train stacked LSTMs to derive nonlinear approximators for these functions. LSTMs can help keep track of the state of the system since they allow modeling continuous systems with slow dynamics. The heating and cooling energy estimated by the LSTMs will be used as a part of the reward function as discussed next.


\subsection{Reward Function}\label{Reward Signal}
The reward function includes two components: (1) the total energy savings for the building expressed as heating and cooling energy savings, and (2) the comfort level achieved. The reward signal at time instant $t$ is by 
\begin{equation}\label{reward}
    r_{t+1}(\mathit{S}_t,\mathit{A}_t) =  \vartheta*Reward_{energy} + (\vartheta)*Reward_{comfort}
\end{equation}
where $\vartheta \in [0,1]$ defines the importance we give to each term. We considered $\vartheta=0.5$ in this work. 

$Reward_{energy}$ is defined in terms of the energy savings achieved with respect to the rule-based controller previously implemented in the building, \textit{i.e.} we reward the RL controller when its actions result in energy savings calculated as the difference between the total heating and cooling energy under the RBC controller actions and the RL controller actions. $Reward_{energy}$ is defined as follows

\begin{multline*}\label{reward energy}
    Reward_{energy}= RBC_{valve,t}*RBC_{heating,t}\\
    -RL_{valve,t}*RL_{heating,t} +RBC_{cooling,t}\\
    -RL_{cooling,t}
\end{multline*}

where the components of this equation are 
\begin{itemize}
    \item $RL_{heating,t}$: The total energy used to heat the air at the heating or preheating coil as well as the VRF system at time-instant $t$ based on the heating set point at the AHU assigned by the RL controller.
    \item $RBC_{heating,t}$: The total energy used to heat the air at the heating or preheating coil as well as the VRF system at time-instant $t$ based on the heating set point at the AHU assigned by the Rule Based Controller(\textit{RBC}).
    \item $RL_{valve,t}$: The on-off state of the heating valve at time-instant $t$ based on the heating set point at the AHU assigned by the RL controller.
    \item $RBC_{valve,t}$: The on-off state of the heating valve at time-instant $t$ based on the heating set point at the AHU assigned by the Rule Based Controller(\textit{RBC}).
    \item $RL_{cooling,t}$: The total energy used to cool the air at the cooling coil as well as the VRF system at time-instant $t$ based on the set point at the AHU assigned by the RL controller.
    \item $RBC_{cooling,t}$: The total energy used to cool the air at the cooling coil as well as the VRF system at time-instant $t$ based on the set point at the AHU assigned by the Rule Based Controller(\textit{RBC}).
\end{itemize}

Here by Rule Based Controller set-point, we refer to the historical set point data that is obtained from the past data on which we shall do our comparison.

The heating and the cooling energy are calculated as a function of the exogenous state variables $\xi_{t+1}$ and $A_t$, as discussed in the previous sub-section. Additionally, we model the behavior of the valve that manipulates the steam flow in the coil of the heating system, This valve shuts off under certain conditions such that the heating energy consumption sharply drops to 0. This hybrid on-off behavior cannot be modeled with an LSTM thus we need to model the valve behavior independently as a on-off switch to decide when to consider the predictions made by the LSTM (only during on). Note that both $RBC_{valve,t}$ and $RL_{valve,t}$ are predicted by using a binary classifier.

The reward for comfort is measured by how close the supply air temperature is to the Average Building Temperature Preference set-point(\textit{avg-stpt}. Let $\Delta_t = abs(\textit{avg-stpt}-rl_{setpoint})$

\[
    Reward_{comfort}
    \begin{cases}
    \frac{1}{\Delta_t+1},& if \Delta_t \leq 10^oF \\
    -\Delta_t, & if \Delta_t > 10^oF
    \end{cases}
\]
The comfort term allows the RL controller to explore in the vicinity of the average building temperature preference to optimize energy. The 1 added to the denominator in case 1 makes the reward bounded. 

The individual reward components are formulated such that a preferred action would provide positive feedback while a negative feedback implies actions which are not preferred. The overall reward is non-sparse so the RL agent would have sufficient heuristic information for moving towards an optimal policy.




\section{Implementation Details}\label{Implementation}
In this section, we describe the implementation of the proposed approach for the optimal control of the system described in the previous section. 

\subsection{Data Collection and Processing}
This process is part of \textit{Step 1} in Figure \ref{fig:approach}. The data was collected over a period of 20 months(July '18 to Feb '20) from the building we were simulating using the \textit{BACNET} system which is a collection of sensor data logging all the relevant variables related to our study. These include the weather variables, the building set points, energy values collected at 5 minute aggregations. We first cleaned the data where we removed the statistical outliers using a 2 standard deviations approach. Next we aggregated the variables at half-an-hour intervals where variables like temperature, humidity were averaged and variables like energy were summed over that interval. Then we scaled the data to a $[0,1]$ interval so that we can learn the different data-driven models and the controller policy. In order to perform the off-line learning as well as the subsequent relearning, we sampled this above data in windows of 3 months(for training) and 1 week(for evaluating).

\subsection{Definition of the environment}

The environment $\mathit{E}$ has to implement the the functions $f_h,f_v,f_c$ as described in Section \ref{Reward Signal} as they will be used to calculate the energy and valve state.

\subsubsection{Heating Energy model}\label{heattrain}
This process is part of \textit{Step 2} in Figure \ref{fig:approach}. The heating energy model is used to calculate the heating energy consumed in state $\mathit{S_{t+1}}$ which results from the action $\mathit{A_t}$ taken in state $\mathit{S_{t}}$. The model for Heating energy $f_h$ is trained using the sequence of variables comprising the states $\mathit{S_{t+1}}$ over the last 3 hours \textit{i.e.} 6 samples considering data samples at 30 minute intervals. The output for the heating energy model is the total historical heating energy over next 30 minute interval.

The heating coils for the building operate in a hybrid mode where the heating valve is shut-off at times. Thus the heating energy goes to zero for that instant. This abrupt change cannot be modeled by a smooth LSTM model. We therefore decided to train our model on contiguous sections where the heating coils were operating. During evaluation phase, the valve($f_v$) model will predict the on/off state of the heating coils. We shall predict the energy consumption only for those instances when the valve model determines the heating coils to be switched on.

The model for $f_h$ is constructed by stacking 6 Fully Feed Forward Neural (\textit{FFN}) Network Layers of 16 units each followed by 2 layers of LSTM with 4 units each. The activation for each layer is \textit{Relu}. The FFN layers are used to generate the rich feature from the input data and the LSTM layers are used to learn the time based correlation. The learning rate is initially 0.001 and is changed according to a linear schedule to ensure faster improvement at the beginning followed by gradual improvements near the optimum so that we don't oscillate around the optima. Mean Square Error on validation data is used to terminate training. The model parameters were found by hyper-parameter tuning via Bayesian Optimization on a  Ray-Tune\cite{liaw2018tune} cluster.

\subsubsection{Valve State model}\label{valvetrain}
This process is also a part of \textit{Step 2} in Figure \ref{fig:approach}. The valve model $f_v$ is used to classify whether the system is switched on or off or equivalently $i.e.$ whether the heating energy is positive or 0. The input to this model is the same as the Heating Energy model. The output is the valve (heating coil) on-off state at the next time instant.

The model for $f_v$ is constructed by stacking 4 Fully Feed Forward Layers of 16 units each followed by 2 layers of LSTM with 8 units each. The activation for each layer is \textit{Relu}. The learning rate, validation data, and the model parameters are similarly chosen as before. The loss used in this case is the binary cross-entropy loss since it is a two-class prediction problem.

\subsubsection{Cooling Energy model}\label{cooltrain}
This process is also part of \textit{Step 2} in Figure \ref{fig:approach}. The cooling energy model is used to calculate the cooling energy consumed in state $\mathit{S_{t+1}}$ when the action $\mathit{A_t}$ is taken in state $\mathit{S_{t}}$. The input to this model is the same as the Heating Energy model. The output of the model is the total historical cooling energy over the next 30 minute interval.

The model for $f_c$ is constructed by stacking 6 Fully Feed Forward Layers of 16 units each followed by 2 layers of LSTM with 8 units each. The activation for each layer is \textit{Relu}. The learning rate, validation data  and the model parameters are chosen in a way similar to the Heating Energy Model.

Once the processes in step 2 are completed we construct the data-driven simulated environment $E$. It receives the control action $\mathit{A}_t$ from the PPO controller and steps from its current state $\mathit{S}$ to the next state $\mathit{S}'$. To calculate $\mathit{S}'$, the weather values for the next state are obtained by simple time-based lookup from the "Weather Data" database. The supply air temperature for the next state is obtained from the "State Transition Model" using equation \ref{sattrans}. The reward $r_{t+1}(\mathit{S},\mathit{A}_t)$ is calculated using equation \ref{reward}. Every time the Environment is called with an action, it will perform this entire process and return the next state $\mathit{S}'$, the reward $r_{t+1}(\mathit{S_t},\mathit{A}_t)$ back to the RL Controller with some additional information on the current episode.

\subsection{PPO Controller}\label{ppotrain}
This process is part of \textit{Step 3} in Figure \ref{fig:approach}. As discussed previously in section \ref{PPO}, the controller will learn two neural networks using the feedback it receives from the environment $E$ in response to its action $\mathit{A}_t$. This action is generated by sampling from the distribution $\mathit{N}(\mu,\sigma)$ which are the outputs of the policy network as shown in the figure. After sampling responses from the environment for a number of times, the collected experiences under the current controller parameters, are used to update the controller network by optimizing $\mathrm{L}^{CLIP}(\theta)$ in equation \ref{Lclip} and the value networks by TD Learning. We repeat this training process until the optimization has converged to a local optima.

The Policy Network model architecture consists of two layers of Fully Feed Forward Layers with 64 units each. The Value Network network structure is identical to the Policy network. The networks are trained on-policy with a learning rate of $0.0025$. Each time the networks were trained over 1e6 steps through the environment. For the environment $\mathit{E}$ this corresponded to approximately 10 episodes for each iteration.

\subsection{Evaluating the energy models, valve models, and the PPO controller}\label{evaluation}
This corresponds to Step 4 in Figure \ref{fig:approach}. Once the energy model, the valve state models, and the controller training have converged we evaluate them on a held out test data for 1 week. The Energy models are evaluated using the Coefficient of variation Root Mean Square Error (\textit{CVRMSE}) 
\begin{equation*}
    CVRMSE = \frac{\sqrt{\sum_{i}(y_{true}-y_{pred})^2}}{\Bar{y_{true}}}
\end{equation*}
where $y_{true}$ and $y_{pred}$ represent the true and the predicted value of the energy, respectively. 

The valve model is evaluated based on its ROC-AUC as the on-off dataset was found to be imbalanced. The controller policy is evaluated by comparing the energy savings for the cooling energy and the heating energy as well as how close the controller set-point for the AHU supply air temperature is to the building average set-point \textit{avg-stpt}.

\subsection{Relearning Schedule}\label{relearning}
Steps 4 and 5 in Figure \ref{fig:approach} are repeated by moving the data collection window forward by 1 week. We observed that having a large overlap over training data between successive iterations helps the model retain previous information and gradually adapt to the changing data.

From the second iteration onward we do not train the data driven LSTM models (\textit{i.e.} $f_h,f_v,f_c$) from scratch. Instead, we use the pre-trained models from the previous iteration to start learning on the new data. For the energy models and valve models we no longer train the FFN layers and only retrain the head layers comprising the LSTMs.  The FFN layers are used to learn the representation from the input data and this learning is likely to stay identical for different data. The LSTM layers, on the other hand, model the trend in the data which must be relearnt due to the distributional shift. Our results show that this training approach saves time with virtually no loss in model performance. We also adapt the pre-trained controller policy according to the changes in the system. This continual learning approach save us time during repeated retraining and allows the data-driven models and the controller adapting to the non-stationarity of the environment.

\section{Results}\label{Results}
In this section we present the performance of our energy models, valve model, and the RL controller over multiple weeks.

\subsection{Relearning Results for Heating Energy Model}

\begin{figure}[!ht]
    \centering
    \includegraphics[width=\columnwidth,height=5.5cm]{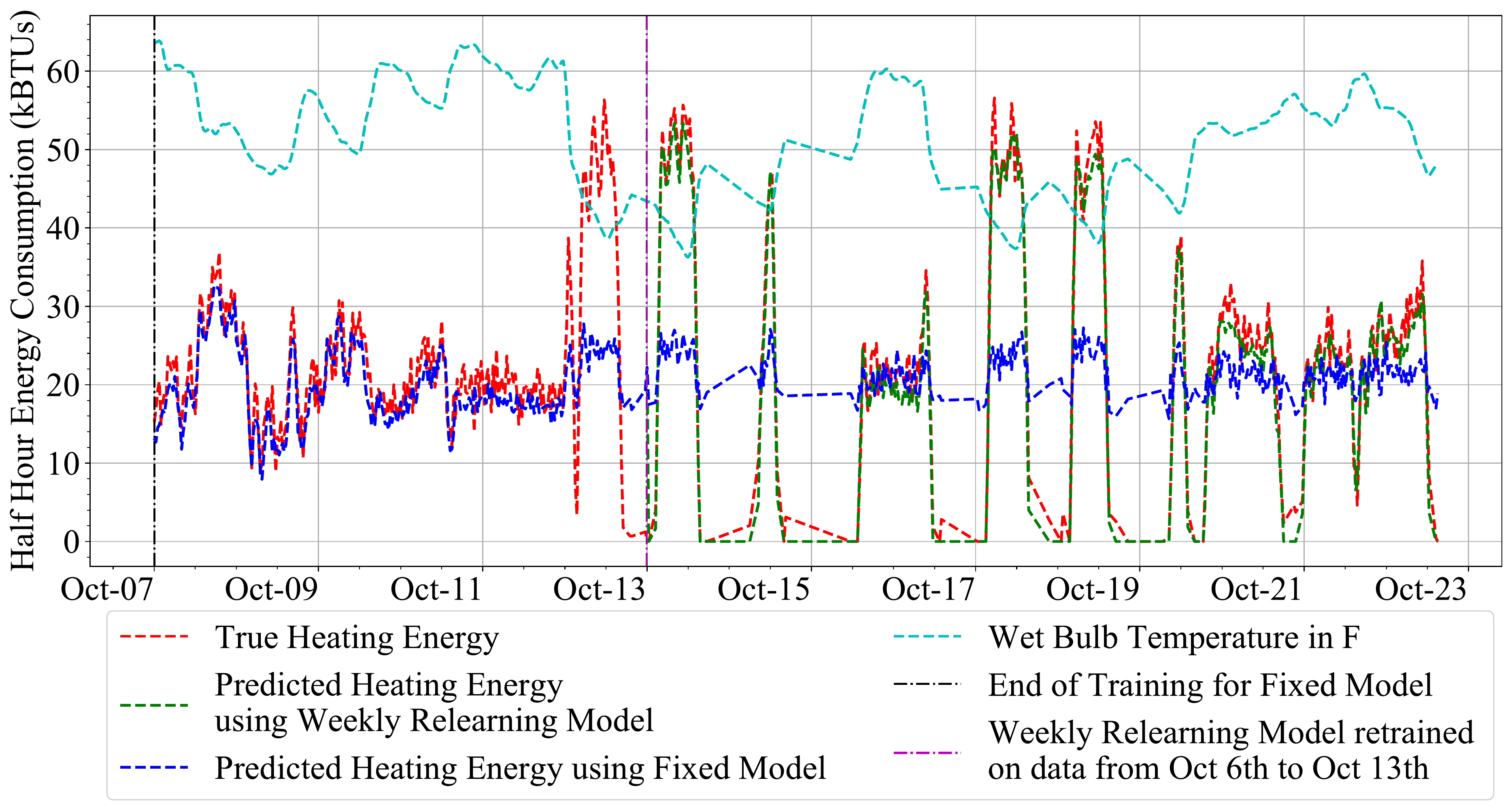}
    \caption{Comparison of true versus predicted Heating Energy for a weekly relearning model and a static/non-relearning model}
    \label{fig:HWE_Plot}
\end{figure}

\begin{figure}[!ht]
    \centering
    \includegraphics[width=\columnwidth]{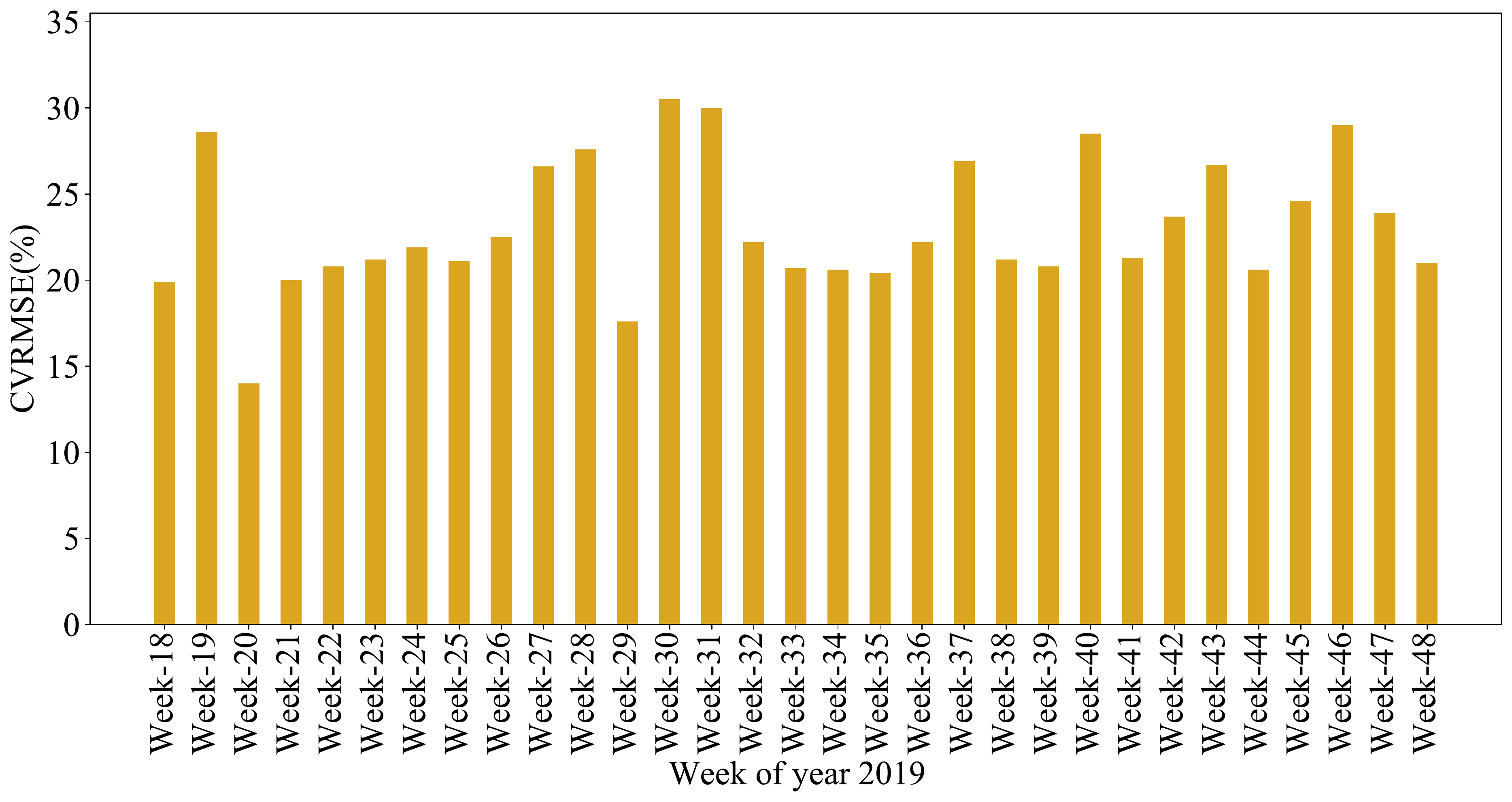}
    \caption{The weekly CVRMSE of the Hot Water Energy Relearning Model for predicting Hot Water Energy consumption at half hour intervals}
    \label{fig:HWE_CVRMSE_Plot}
\end{figure}

Figure \ref{fig:HWE_Plot} shows the heating energy prediction on a subset of the data from October 7th to 23rd. We selected this time period because the effects of the non-stationarity in the data can be appreciated. We compare the prediction of a fixed model, which is not updated after October 7th, with a model which is retrained by including the new week's data from 7th to the 13th.
The figure demonstrates the necessity of relearning the heating energy model at regular intervals. After the October 12th, the AHU switches from using the reheating to the preheating coil due to colder weather as indicated by the wet bulb temperature. This causes the heating energy consumption to change abruptly. The model which is not updated after October 7th is not able to learn this behavior and keeps predicting similar behavior as before. On the other hand, the weekly relearning model behavior starts degrading but once it is relearned using the data from Oct 7th to the 13th, it can capture the changing behavior quickly using a small section of similar data in its training set. The overall CVRMSE for the relearning energy model is shown in Figure \ref{fig:HWE_CVRMSE_Plot}. For majority of the weeks, the CVRMSE is below $30\%$ which is accepted according to ASHRAE guidelines for energy prediction at half hour intervals

\subsection{Relearning Results for Cooling Energy Model}

\begin{figure}[!ht]
    \centering
    \includegraphics[width=\columnwidth,height=5.5cm]{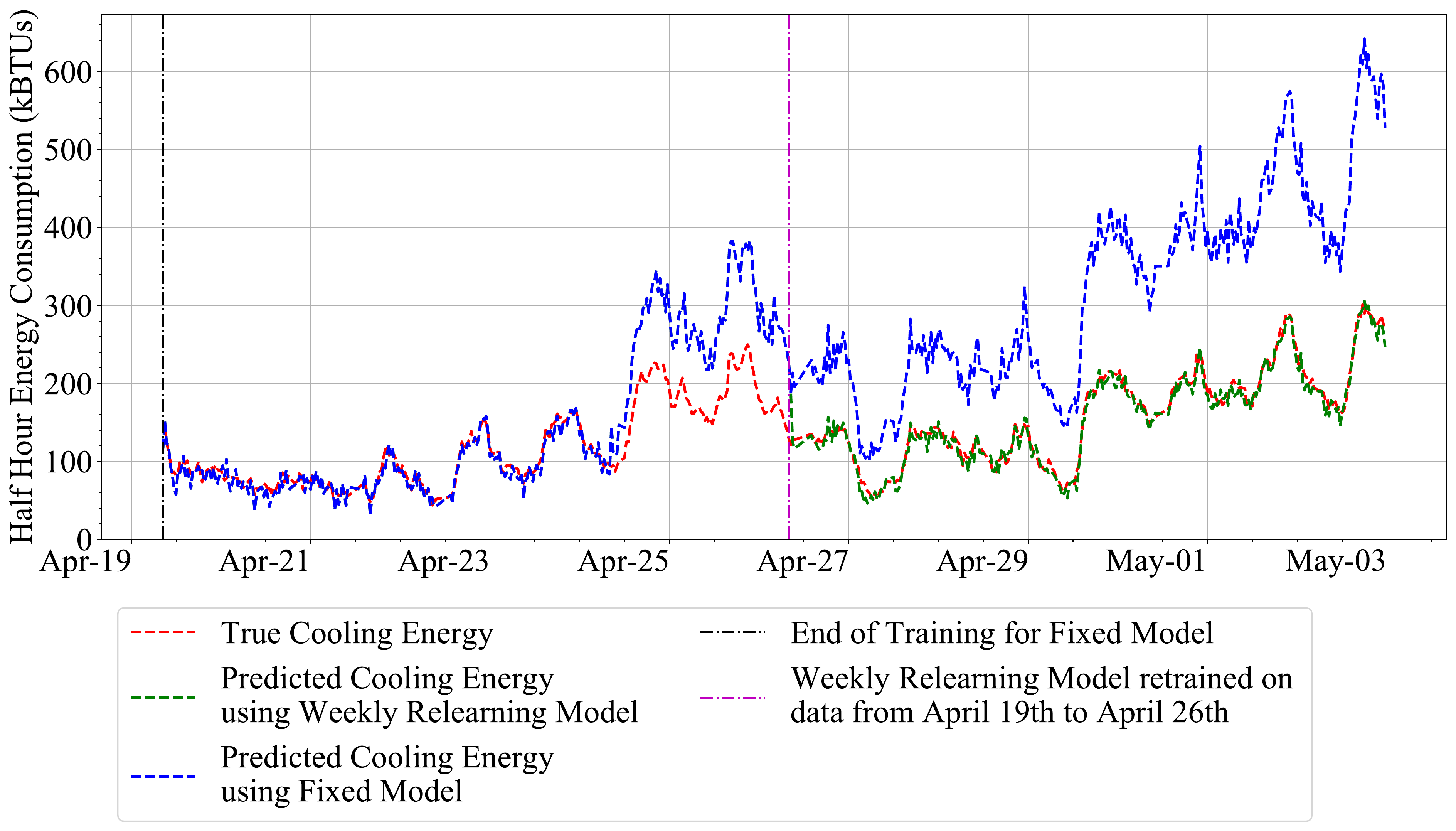}
    \caption{Comparison of true versus predicted Cooling Energy for a weekly relearning model and a static/non-relearning model}
    \label{fig:CWE_Plot}
\end{figure}

\begin{figure}[!ht]
    \centering
    \includegraphics[width=\columnwidth]{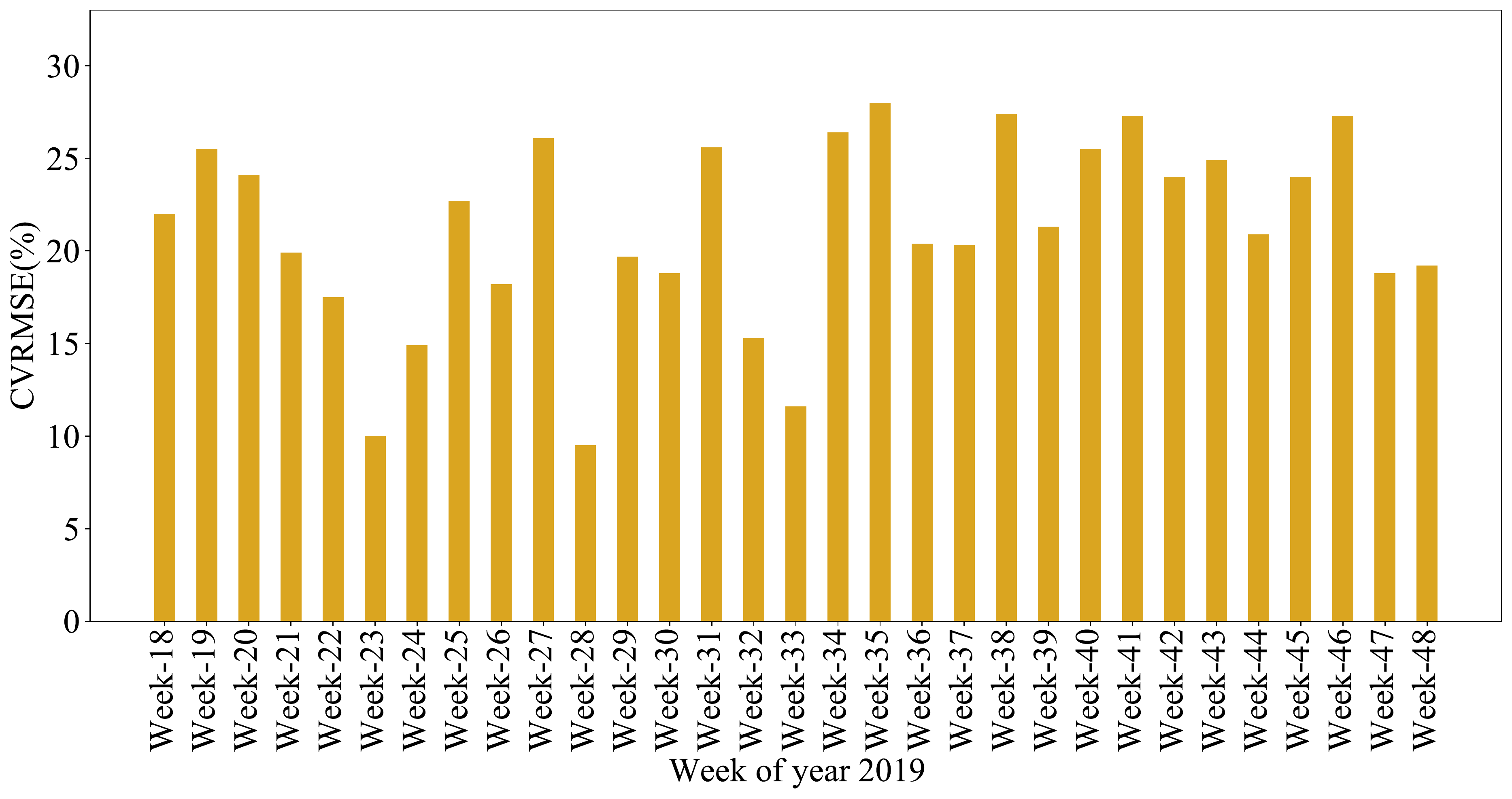}
    \caption{The weekly CVRMSE of the Cooling Energy Relearning Model for predicting Cooling Energy consumption at half hour intervals}
    \label{fig:CWE_CVRMSE_Plot}
\end{figure}

Figure \ref{fig:CWE_Plot} shows the plots for predicting the Cooling energy Energy over a span of two weeks. We also include the the energy prediction from a fixed model. Starting from 25th April, both the Fixed and Relearning model for Cooling Energy  predictions start degrading as they start following an increasing trend while the actual trend is downward and this behavior is expected while learning on non-stationary data. But the Relearning Cooling Energy model is retrained using the data from April 19th to April 26th at the end of the week corresponding to 26th April. Thus its predictions tend to be better than a fixed model for the next week whose predictions degrade as the week progresses.
The overall CVRMSE for the relearning energy model is shown in Figure \ref{fig:CWE_CVRMSE_Plot}. For all the weeks, the CVRMSE is below $30\%$ which is accepted according to ASHRAE guidelines for energy prediction at half hour intervals

\subsection{Prediction of the Heating Valve status}
\begin{figure}
    \centering
    \includegraphics[width=\columnwidth]{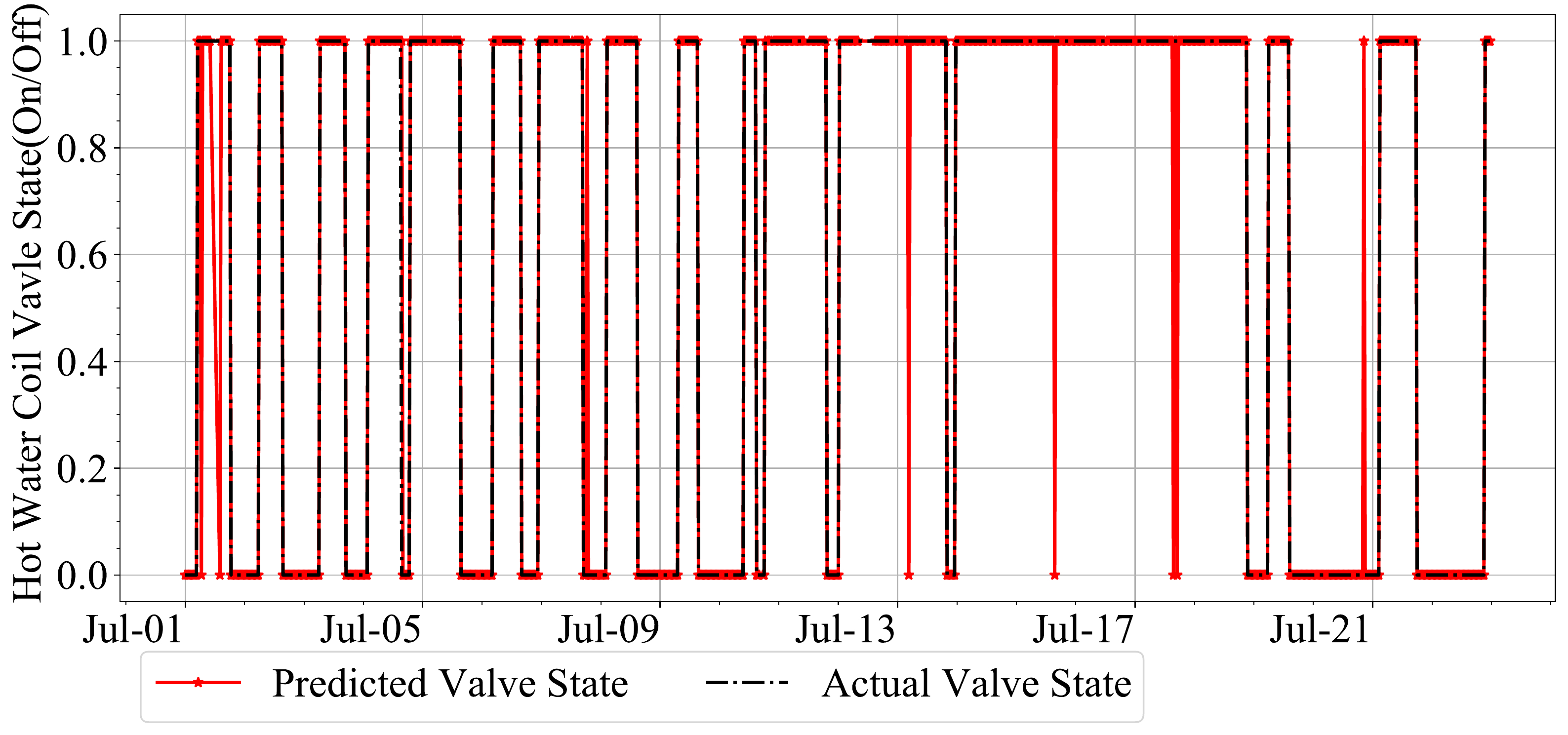}
    \caption{Comparing True versus Predicted Hot Water Valve State behavior}
    \label{fig:VLV_plot}
\end{figure}

\begin{figure}[!ht]
    \centering
    \includegraphics[width=\columnwidth]{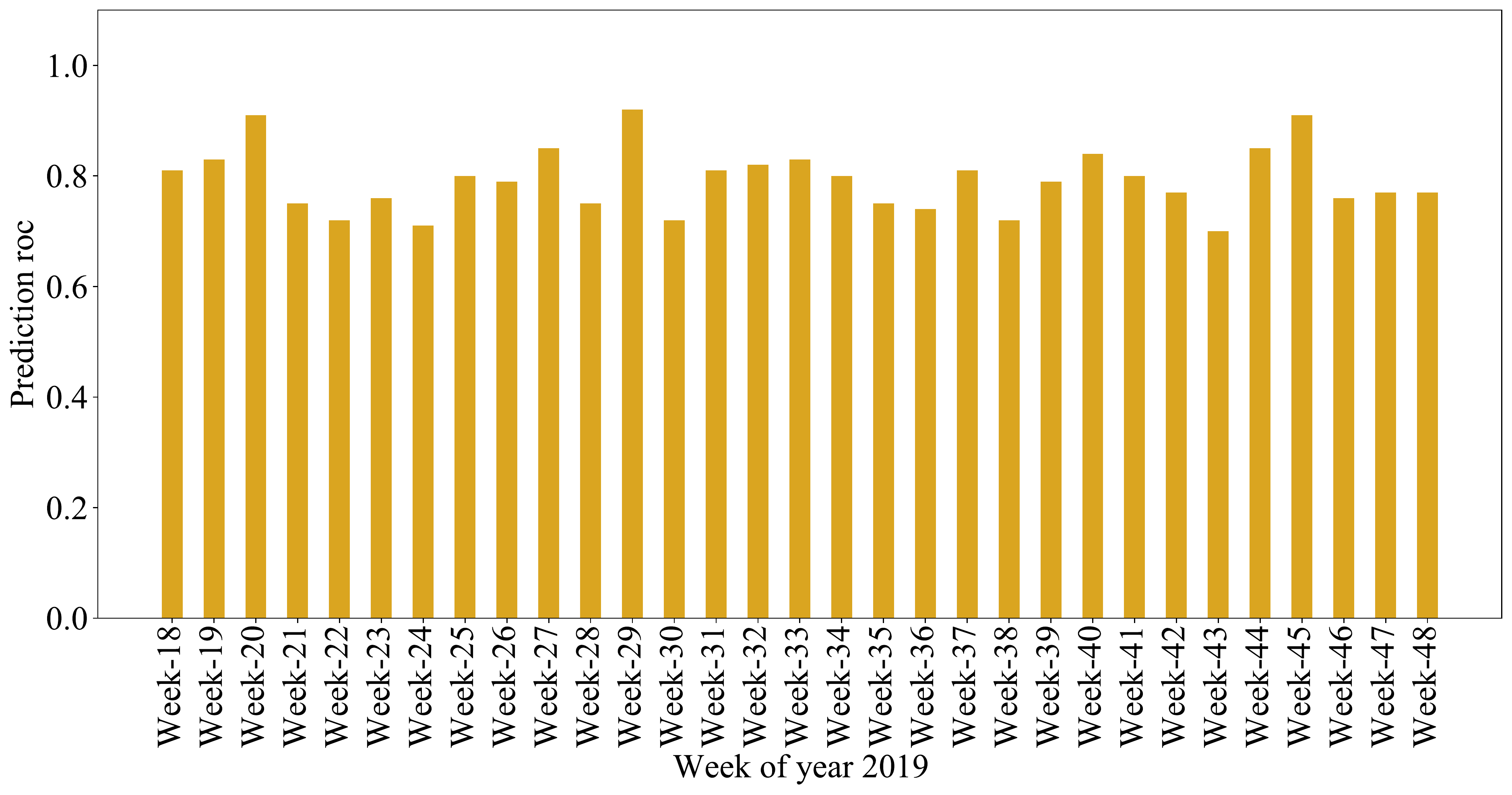}
    \caption{Hot Water Valve State Prediction model ROC AUC evaluated over multiple weeks}
    \label{fig:VLV_Roc}
\end{figure}

Figure \ref{fig:VLV_Roc} shows the Area Under the Receiver Operating Characteristics (ROC AUC) for the model predicting the valve status(on/off). We also show the actual and predicted valve state for around one month in Figure \ref{fig:VLV_plot}. Overall, the relearning valve model is able to accurately predict the valve behavior.

\subsection{Training Episode Reward}
\begin{figure}
    \centering
    \includegraphics[width=\columnwidth]{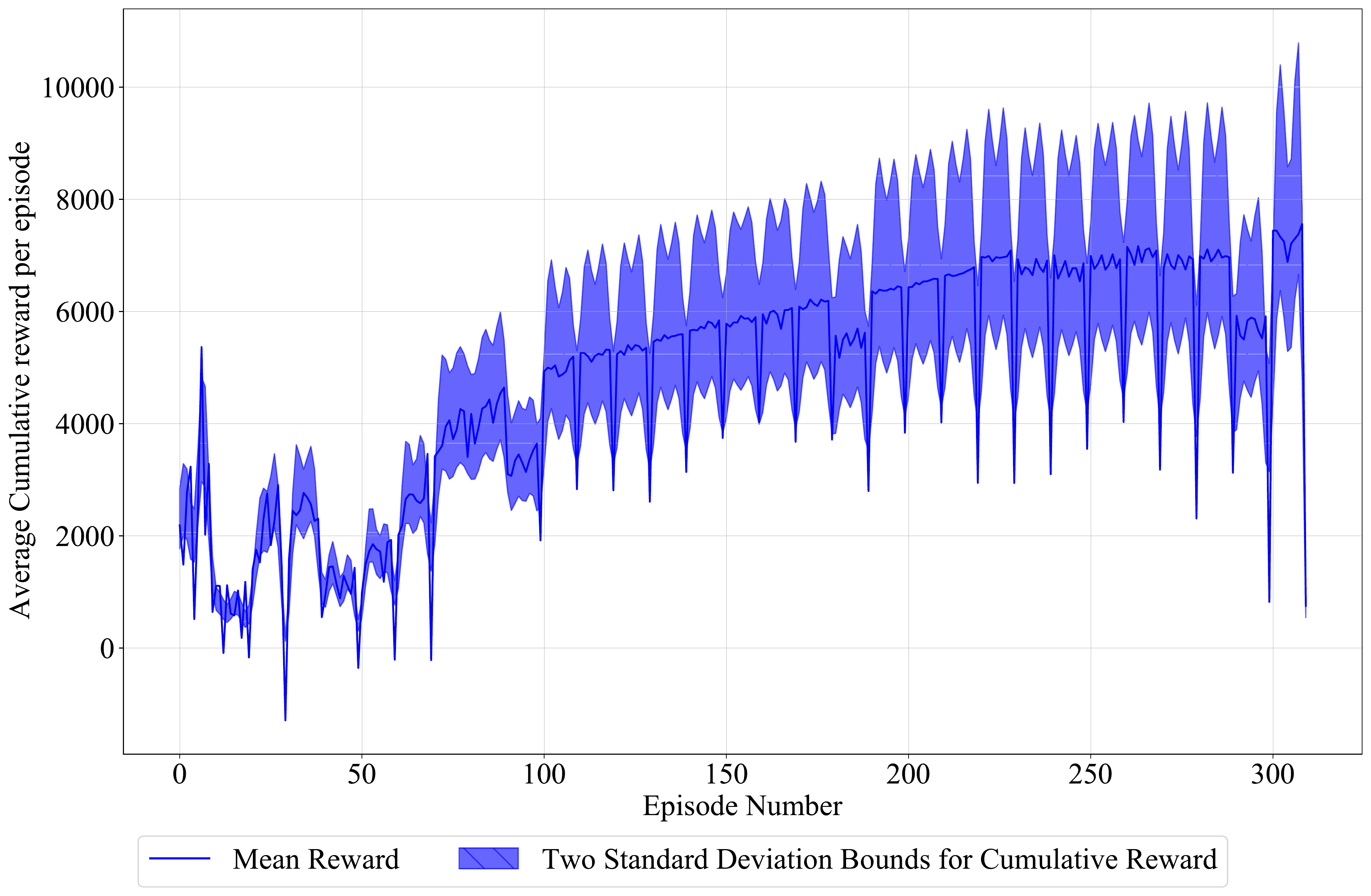}
    \caption{Average Cumulative Reward Obtained across each episode trained across 10 environments in parallel}
    \label{fig:AverageReward}
\end{figure}

We trained the PPO controller on the environment $\mathit{E}$ every week to adjust to the shift in the data. The cumulative reward metric from equation \textit{J} is used to asses the improvement in controller performance over the number of week. We observed that even though the controller is able to achieve good results after training over a couple of weeks of data, it still keeps improving as weeks progresses. The cumulative reward metric is plotted in Figure \ref{fig:AverageReward}. The occasional drops in the average reward are due to changing environment conditions as training progresses.

\subsection{Cooling Energy Performance}

\begin{figure}[!ht]
    \centering
    \includegraphics[width=\columnwidth,height=5.5cm]{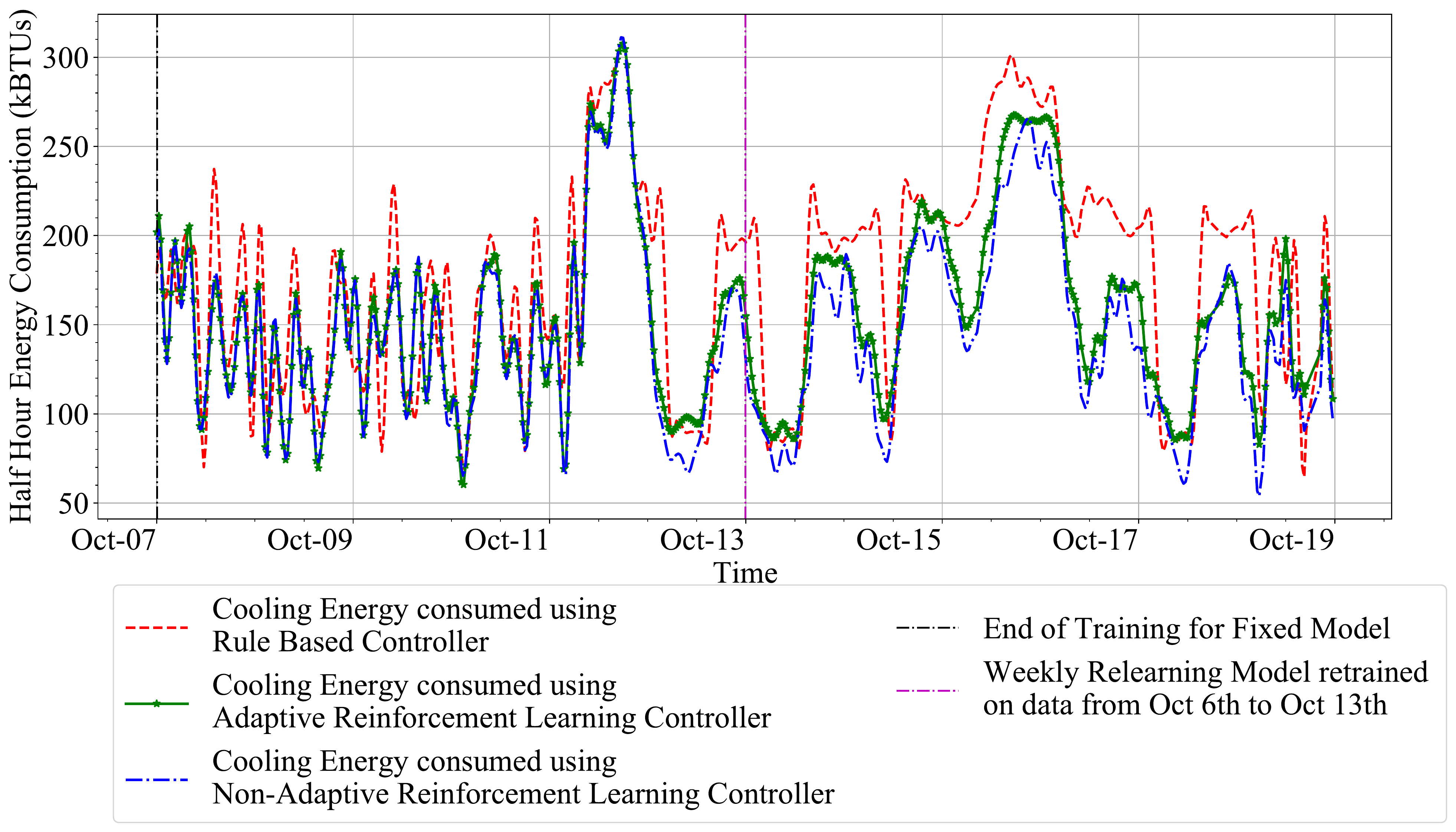}
    \caption{Plot of Cooling Energy Consumed for actions based on RBC, Adaptive RL controller and Static RL Controller}
    \label{fig:cwe_rl_plot}
\end{figure}

We compared the cooling energy performance of both the adaptive reinforcement learning controller and a static reinforcement learning controller against a rule based controller. A plots comparing the cooling energy consumed over a certain part of the evaluation period is shown in figure \ref{fig:cwe_rl_plot}. We are displaying this part of the time-line because it will be significant in understanding why relearning is important. When we calculate the energy savings for each RL controller, the static RL controller had slightly higher cooling energy savings because the last version of it was trained during warmer weather and it tends to keep the building cooler. But when the outside temperature drops, the static controller action does not heat the system too much resulting in the VRF systems starting to heat the building which consume higher energy. The cooling energy savings over the period shown in figure \ref{fig:cwe_rl_plot} was $9.3\%$ for the Adaptive Controller and $ 11.2\%$ for the Static controller. The average weekly cooling energy savings over the entire evaluation period of 31 weeks was $12.61\%(5.73\%)$ or $188.53(18.153)$ kBTUs for the Adaptive Controller versus $12.81\%(8.22\%)$ or $191.21(23.009)$ kBTUs  for the Non-Adaptive/Static Controller.

\subsection{Heating Energy Performance}
\begin{figure}[!ht]
    \centering
    \includegraphics[width=\columnwidth,height=5.5cm]{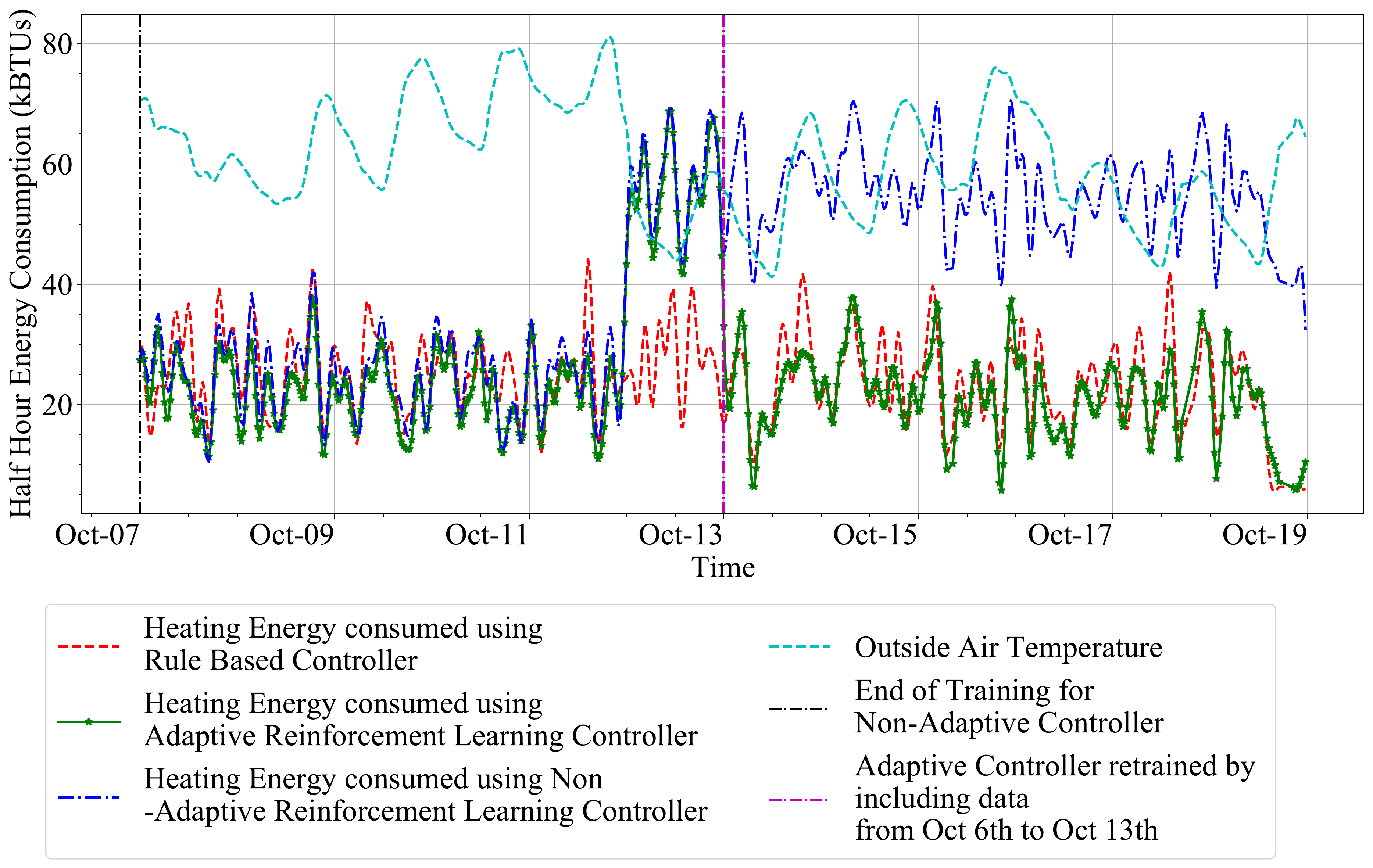}
    \caption{Plot of Heating Energy Consumed for actions based on RBC, Adaptive RL controller and Static RL Controller}
    \label{fig:hwe_rl_plot}
\end{figure}

Similarly, we compared the heating energy performance of an adaptive and static controller over the same timeline as shown in figure \ref{fig:hwe_rl_plot}. This plot shows the severe issue of over-cooling that can occur in the building when controller is not updated regularly, Due to lower action set point of the static controller, the total heating energy consumption for the building goes up over the entire period of cool weather. The heating energy savings over the period shown in figure \ref{fig:cwe_rl_plot} was $6.4\%$ for the Adaptive Controller while the Static controller increased the energy consumption by $65\%$. The average weekly heating energy savings over the entire evaluation period of 31 weeks was $7.19\%(2.188\%)$ or $112.19(13.91)$ kBTUs for the Adaptive Controller whereas the Non-Adaptive/Static Controller increased the energy consumption by $54.88\%(32.66\%)$ or $161.08(18.211)$ kBTUs.

\begin{figure}[!ht]
    \centering
    \includegraphics[width=\columnwidth,height=5.5cm]{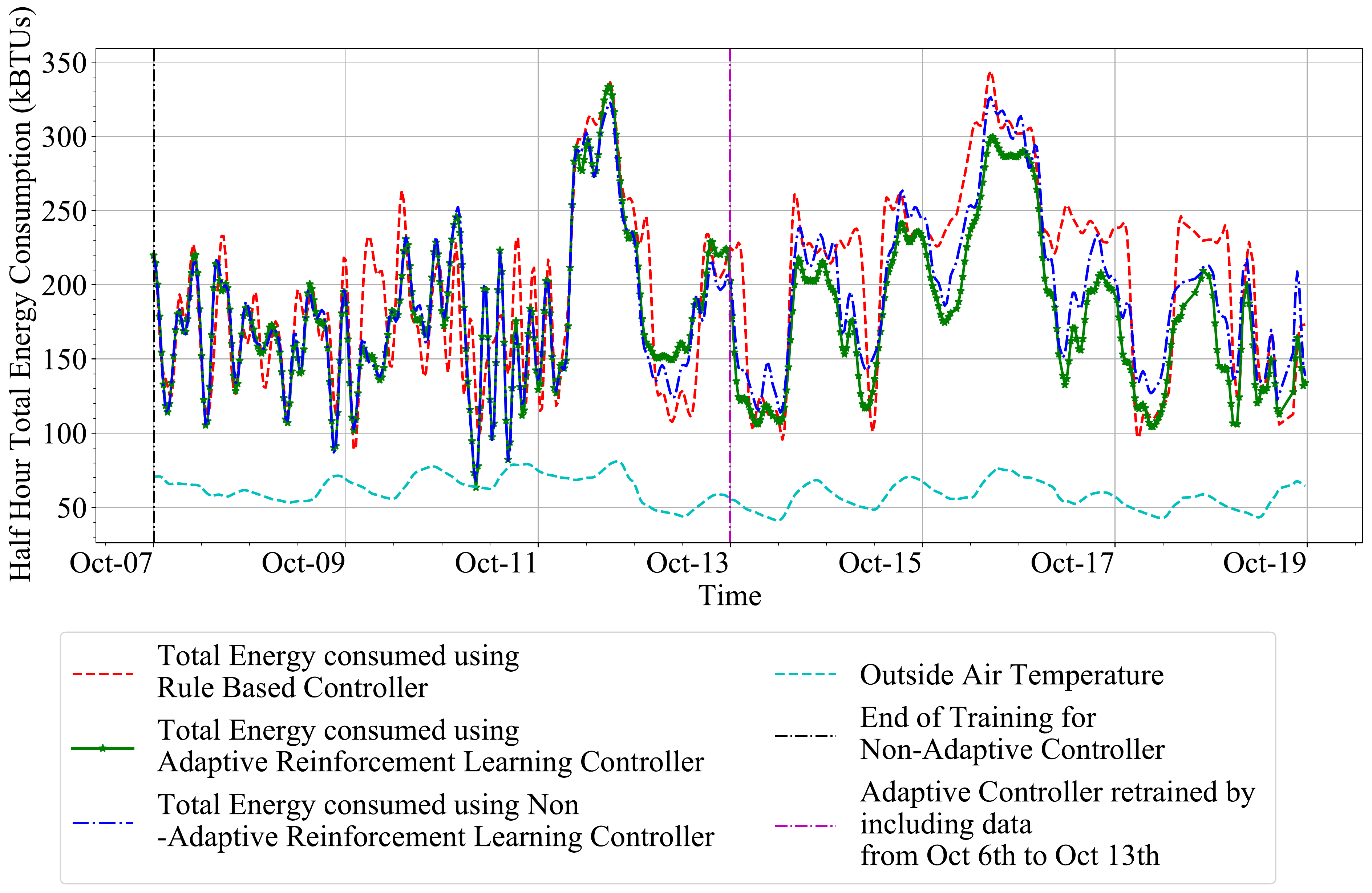}
    \caption{Plot of Total Energy Consumed for actions based on RBC, Adaptive RL controller and Static RL Controller}
    \label{fig:total_rl_plot}
\end{figure}

The sum total of the heating and cooling energy consumption under the historical rule based controller, the adaptive controller and the non-adaptive controller is shown in figure \ref{fig:total_rl_plot}. The adaptive controller consistently saves more energy than the non-adaptive controller. Overall the adaptive controller was able to save 300.72 kBTUs each week on average whereas the static controller was able to save only 30.03 kBTUs.

\subsection{Control Actions}
\begin{figure}[!ht]
    \centering
    \includegraphics[width=\columnwidth,height=5.5cm]{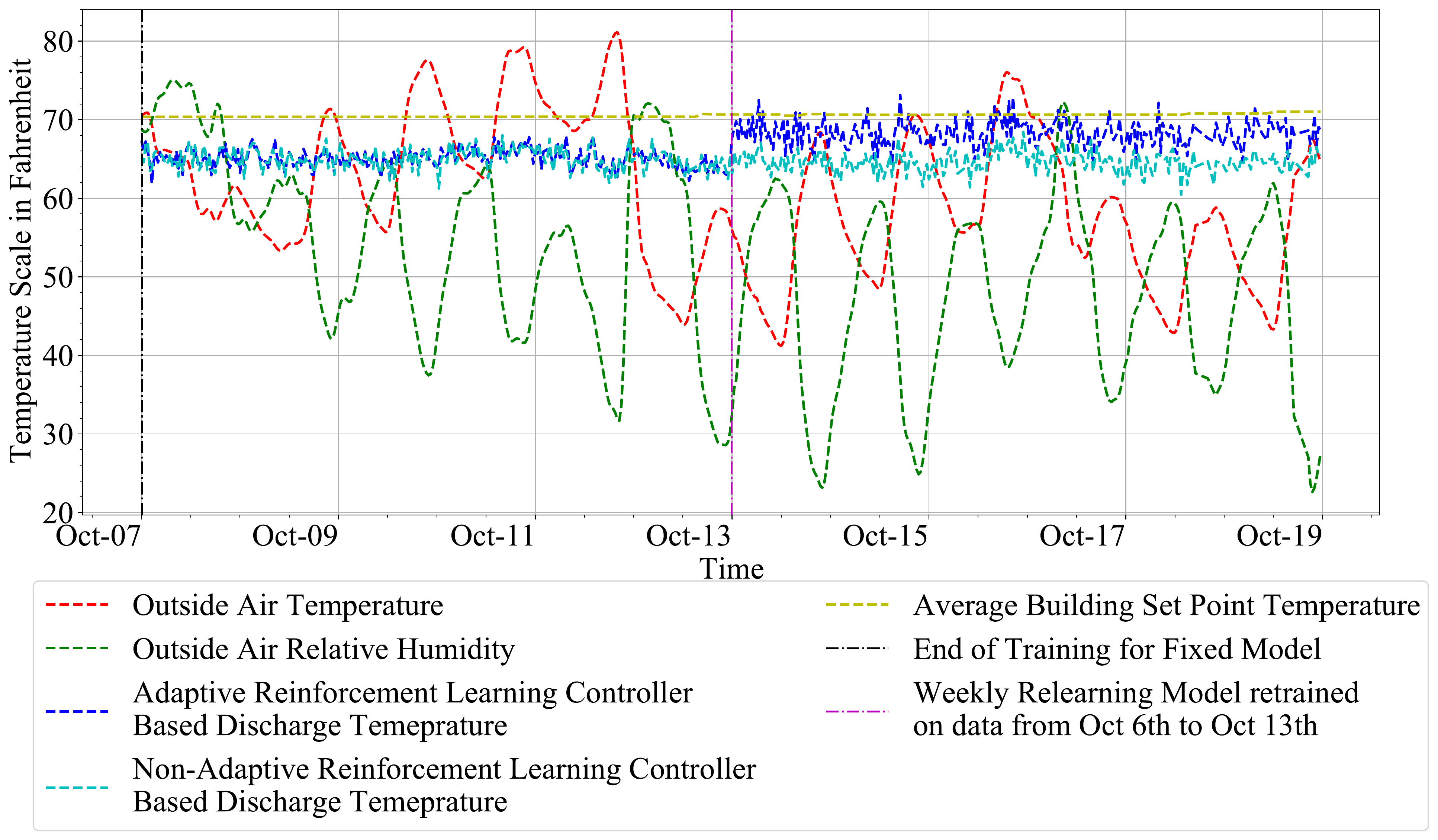}
    \caption{Plot of Supply Air setpoint(\textit{sat}) based on actions chosen by the Adaptive RL controller versus Static RL Controller}
    \label{fig:var_rl_plot}
\end{figure}

Here we show why the overall energy consumption of the building went up when we use a static controller. We plot the Discharge/Supply Air Temperature set-point resulting from the actions of both the adaptive and static controller along with outside air temperature and relative humidity in Figure \ref{fig:var_rl_plot}. On October 12th, the outside temperature goes down and both the adaptive and static controller fail to improve building comfort condition. After October 13th , the adaptive controller is re-trained by considering the last weeks data where it encounters environments states with lower outside air temperatures as subsequently it adapts to those conditions. For the remaining of the time period analyzed the adaptive controller keeps the Supply Air Temperature set-point closer to the comfort conditions required by the occupants. 

\section*{Conclusions}\label{Conclusion}
We demonstrated the effectiveness of including retraining in a data-driven reinforcement learning framework.

It may be argued that our reward is only improving against a baseline Rule Based Controller. The truth is that we can only compare against controllers which select reasonable actions within the distribution of the data on which the data driven models were trained. If we were to learn our reinforcement learning controller without any comparison during training, the exploratory behavior of reinforcement learning methods may have found even better control actions. But as we are using data-driven models, it is highly likely that the actions chosen by the controller might lead the data-driven models to extrapolate results and introduce Out of Distribution Error. By comparing against a rule based controller and constraining actions from veering too far from the current actions, we might leave some savings but we can ensure that the data-driven models used in the environment are not leading us to spurious results by extrapolating.


\bibliographystyle{apacite}
\bibliography{PHM2016_Latex_Template}

\end{document}